# Leveraging Unstructured Data to Detect Emerging Reliability Issues

Deovrat Kakde, SAS Institute
Arin Chaudhuri, SAS Institute

## SUMMARY & CONCLUSIONS

Unstructured data refers to information that does not have a predefined data model or is not organized in a pre-defined manner [1]. Loosely speaking, unstructured data refers to text data that is generated by humans. In after-sales service businesses, there are two main sources of unstructured data: customer complaints, which generally describe symptoms, and technician comments, which outline diagnostics and treatment information. A legitimate customer complaint can eventually be tracked to a failure or a claim. However, there is a delay between the time of a customer complaint and the time of a failure or a claim. A proactive strategy aimed at analyzing customer complaints for symptoms can help service providers detect reliability problems in advance and initiate corrective actions such as recalls.

This paper introduces essential text mining concepts in the context of reliability analysis and a method to detect emerging reliability issues. The application of the method is illustrated using a case study.

## 1  INTRODUCTION

Customer and Technician comments are rich sources of symptoms and diagnostic information. Traditionally, customer complaints are recorded in the service provider's database or with government organizations that are responsible for product safety oversight such as the National Highway Transportation Safety Administration (NHTSA). In today's digital world, customer complaints can be found on social media sites such as Facebook and Twitter.

The number of data sources as well as the speed and scale at which the customer complaint or feedback data is generated can be both a blessing and a curse. A manufacturer or a service provider who can harness this data and identify complaint topics can gain a competitive advantage. The remainder of this paper will explain and showcase how text mining techniques can be used to analyze the vast body of customer complaints data for detecting emerging reliability issues. Section 2 introduces the basics of text mining. Section 3 outlines how text mining can be used to identify themes or topics in customer complaints. Section 4 provides a method for identifying emerging reliability issues once a topic is established. An example that uses actual customer complaints is provided in section 5. It is followed by the conclusions.

## 2  BASICS OF TEXT MINING

Text mining is an umbrella term for describing technologies that analyze and process unstructured data. Text mining can mean different things to different analysts [2]. But any text mining technology first involves converting text into numbers and then applying a relevant algorithm to solve the business problem at hand. Input to the text mining algorithm consists of terms, documents, and a corpus. In the context of this paper, each complaint is a document, a document is a collection of terms, and a collection of documents is called a corpus.

The vector space model is commonly used for numeric representation of a document. In this model, each document is a vector, and the elements of the vector indicate the occurrence of a term within a document. A term count or a Boolean value indicates the presence or absence of a word. A document can also be represented as a weighted vector. The weighted vector entries are computed using a weighing approach such as the TF-IDF (Term Frequency-Inverse Document Frequency) [2]. The entries of the weighted vector reflect how important a word is to a document in the corpus.

The corpus is a collection of multiple document vectors and is represented using a term-document matrix. The vector space model ignores the order of words. This model is called the "bag-of-words" approach. The bag-of-words approach works fine for key text mining tasks such as clustering and topic identification. Figure 1 illustrates these key concepts.

*Figure 1 – Term, Document, and Corpus*

Complaint # 1 (Document #1):

Steering of my car locks while turning

Complaint # 2 (Document #2):

Vibrations at steering when car is running at high speed.

A two-document corpus and its equivalent term-document matrix

| | Document 1 | Document 2 |
|---|---|---|
| steering | 1 | 1 |
| of | 1 | 0 |
| my | 1 | 0 |
| car | 1 | 1 |
| locks | 1 | 0 |
| while | 1 | 0 |
| turning | 1 | 0 |
| vibration | 0 | 1 |
| at | 0 | 1 |
| when | 0 | 1 |
| is | 0 | 1 |
| running | 0 | 1 |
| at | 0 | 1 |
| high | 0 | 1 |
| speed | 0 | 1 |

The size of the term-document matrix can grow very fast as the number of documents increases. Additionally, since each document contains only a few terms from the overall corpus, the term-document matrix is also very sparse. Hence, it is important to reduce the dimension of this matrix so that subsequent algorithms work faster and any noise in the data is reduced. The dimension reduction can be achieved by eliminating terms or documents or by using mathematical techniques such as Singular Value Decomposition [3]. These two techniques are outlined below.

2.1 *Dimension Reduction using Term Elimination:*

The number of terms can be reduced by adopting the one or more of the following techniques:

- Stop List: Create a predefined list of terms that are not important for solving the business problem at hand. For the corpus in Figure 1, stop list candidates would be terms such as *my, at, is, when,* and *while*.

- Start List: This is the list of important words. Any term that is not part of the start list is not included in the analysis.

- Stemming: This technique maps a word to its root. For example, words such as *driven* and *driving* are mapped to their root, which is *drive*. There are different algorithms for stemming. Most stemming algorithms in English are based on Porter's stemming algorithm [4].

- Spelling Corrections: Each misspelled word that is not corrected or mapped to its correct spelling forms a separate row in the term document matrix and thus increases the matrix dimension. The spelling correction is generally done using fuzzy matching algorithms. SOUNDEX is one such algorithm [5].

- Remove terms that occur in a single document or less than a set threshold number of documents.

- Remove numerals if they are not of interest to the analysis.

2.2 *Dimension Reduction Using Singular Value Decomposition (SVD):*

SVD is a matrix factorization method. It is used to approximate any high dimensional, term-document matrix by using a lower dimensional matrix [2]. SVD is similar to Principal Component Analysis. Mathematically the SVD of matrix A with rank r is

$$A = U \Sigma V^T \qquad (1)$$

where *U* and *V* are matrices with orthogonal columns. $\Sigma$ is a diagonal matrix with positive singular value entries. The matrix *A* can be approximated by best rank *k* matrix $A_k$ where $k \leq r$.

$$A \cong A_k = U_k \Sigma_k V_k^T \qquad (2)$$

where $U_k$ and $V_k$ are created using first *k* columns of *U* and V $\Sigma_k$ is the *kxk* diagonal matrix with top *k* singular values.

$$\|A - A_k\|_2 = \min_{\text{rank } B \leq k} \|A - B\|_2 \qquad (3)$$

### 3 IDENTIFYING DOCUMENT TOPICS

A corpus or a document can contain multiple topics or themes. A corpus of customer complaints generally consists of topics that are related to product parts or subsystems and descriptions about what went wrong or the symptoms. Complaints also contain information about how the customer was using the product. Consider the following two customer complaints:

Complaint 1:

*I have a 2008 Speed Machine Savoy - the power steering is extremely difficult at times- the light comes on and off- showing a problem is present- I am unable to turn the car normally- making me use two hands-also, there is a rattle in the back- left- very noticeable- to all in vehicle.*

Complaint 2:

*I purchased the car literally off the showroom floor. One day and 38 miles later it stalled out on my wife on the way to church. Two weeks later I got it back. It turned out to be a problem in the accelerator pedal. The dealer had to finally take one from another car to help solve my problem. Two days later I now have a problem with the rear "third eye" brake light. It will not go off. I am returning the vehicle on Monday and will wait to see what this problem is. I know the speed machine savoy is new but I shouldn't be having these kind of problems.*

A quick look at the complaints indicates presence of multiple topics in each complaint. The first complaint is about a 2008 Speed Machine Savoy's power steering, turning problems, and a rattling sound heard at the time of event. The second complaint is about the Speed Machine Savoy's accelerator pedal, stalling, and brake light. The complaint also provides some additional details such as mileage (38 miles) and driver (owner's wife) at the time of the incident. Both complaints were about the Speed Machine Savoy, and each complaint had multiple topics in it. These topics are important for the manufacturer because they shed light on the reliability of the product in the field. An effective text mining algorithm should be able to identify different topics in the document corpus.

There are different techniques to identify topics in a corpus [2]. For the case study in the current paper, SAS® Enterprise Miner™ was used for topic identification. The algorithm started with the basic SVD concept vectors and then applied a rotation transformation to try to produce concepts that can be more easily interpreted.

## 4 IDENTIFICATION OF EMERGING RELIABILITY ISSUES

There is a significant amount of literature on using event count data to detect emerging reliability issues. Use of unstructured data to identify emerging issues in product reliability is a relatively new field. However, there is a sufficient amount of research where unstructured data is successfully used to detect emerging issues in the domain of public health. [6] used patient's symptoms that were noted at emergency room admission to detect the onset of an epidemic. This paper uses an approach similar to [6], but applied to the customer complaints.

The identification of useful topics from customer complaints that are indicative of reliability problems or product malfunction issues is the most challenging part. It requires domain knowledge to identify which component, part is problematic as well as which symptoms warrant investigation. Also, customer complaints are generally very verbose, and there is usually a lot of information in complaints that is not useful from the analysis perspective. A good topic should identify the malfunctioning part or assembly as well as shed some light on the symptoms. The dimension reduction techniques such as stop list, start list, stemming, and spelling correction are very useful in eliminating the noise from the complaints and identifying topics of interest.

Any algorithm to identify emerging issues should generate alerts for known problems, as well as any previously unknown problems. In the case of customer complaints, the algorithm should detect, at a minimum, topics that can identify the component, subsystems, and symptoms that were previously unknown.

To identify emerging reliability issues using unstructured data involves first using a text mining algorithm to identify important relevant topics. Then statistical techniques are used to alert for significant shifts in topic frequency. The methodology of detecting shifts is based on the expectation-based Poisson statistic as outlined in [7]. The following hypotheses were tested using this methodology:

$H_0$: $C_{jt}$ the count of documents for a topic j (j=1 to k) in a time period t follows Poisson ($b_j t$) where $b_j$ is the baseline topic rate per document per time period.

$H_1$: $C_{jt}$ the count of documents for a topic j (j=1 to k) in a time period t follows Poisson ($q b_j t$) where $b_j$ is the baseline topic rate per document per time period, for some q >1

$B_{jt}$: Expected count of documents for topic j (j=1 to k) in time period t

n: number of time intervals for calculating $C_{jt}$. At the end of nth interval, the expectation-based Poisson statistic $F_{jn}$ is calculated.

$$C_j = \sum_{t=1}^{n}(C_{jt}) \qquad (4)$$

$$B_j = \sum_{t=1}^{n}(B_{jt}) \qquad (5)$$

$$F_{jn} = \begin{cases} \left(C_j \log \frac{C_j}{B_j} + B_j - C_j\right); & C_j > B_j \\ 0; & C_j \leq B_j \end{cases} \qquad (6)$$

Specific steps are as follows:
- Establish existing topics (k) and topic rate per product per time period ($b_j$, j=1 to k) using historic comments data. The actual document frequency is divided by expected population of products in field to arrive at the topic rate per product, per time period.
- $b_0$=min($b_j$, j=1 to k) is the baseline topic rate for new topics.
- In every new time period, calculate the actual count $C_{jt}$, j=1 to k
- In every $n^{th}$ time period, compute the Poisson log likelihood ratio $F_{jn}$
- The topics with $F_{jn}$ greater than 1, are emerging topics.

The text mining algorithm used for topic identification, assigns one document to multiple topics. If topics are indicative of a part and/or symptoms, then further analysis to identify which topics appear together can be beneficial to understand parts malfunctioning together.

## 5 CASE STUDY

The actual customer complaints were analyzed to identify emerging reliability issues using the methodology described in sections 2 through 4. The complaints were for a specific model of an automobile manufacturer. There were total 5,259 complaints from 1992 to 2014. The first 36 months of data, where each month had at minimum one complaint, was used to identify the existing topic and establish its baseline rate. Analysis was performed using the text mining functionality available in SAS® Enterprise Miner™ version 12.1. In SAS® Enterprise Miner™, the data mining process is driven by a process flow diagram that can be created by dragging nodes from the application toolbar and dropping them in a diagram workspace [7]. The current analysis was performed using the nodes for text import, text parsing, text filter, and text topics. Figure 2 illustrates the process flow diagram. The SAS Code node was used to query the data for a specific time period.

*Figure 2 – Process Flow Diagram*

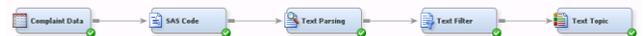

A stop list and a synonym list was specified under the text parsing node. Terms that appear in fewer than four documents were ignored in the analysis. A stop list was carefully constructed so that terms that are not relevant for the analysis are excluded.

The Text Topics node, in addition to finding new topics, enables users to specify a list of custom topics. Users can define topics of interest and force the text mining node to identify documents that belong to these custom topics. A set of custom topics was created based on domain knowledge. Table 1 shows the list of custom topics used in the analysis. As seen in table 1 these topics indicate the important subsystems in automobiles and possible symptoms these automobiles can exhibit. The weight associated with a term indicates the relative importance of the term to the topic. A weight of 1 is highest importance and weight of 0 is the lowest.

*Table 1 – Custom Topics*

| Topic | Terms | Weight |
|---|---|---|
| Vehicle Speed Control | Accelerator | 0.9 |
| Vehicle Speed Control | Gas Pedal | 0.9 |
| Vehicle Speed Control | Cruise Control | 0.8 |
| Vehicle Speed Control | Vehicle Speed Control | 0.8 |
| Accident | Accident | 0.9 |
| Accident | Crash | 0.9 |
| Accident | Collision | 0.9 |

The analysis was first run using the 36 months of data. After analysis completion, the Text Topics node assigned documents to the custom topics as well as newly discovered topics. Each of the new topics was examined for relevance. If the topic was of interest, then it was converted into a custom topic, so that the Text Topic node could assign documents available in the next time period to this topic. One of the topics discovered when analyzing the first 36 months data was failure of power steering with locking. The key terms that described this topic were *power, steer, lock, indicator, light,* and *illuminate*. This discovered topic consisted of a subsystem, Power Steering in this case, and symptoms such as locking and indicator light illumination.

After analyzing the first 36 months of customer complaints data, the subsequent data was analyzed in three-month increments. For each month, the $C_{jt}$ was compared against the baseline count $B_{jt}$ to compute the Poisson log likelihood ratio $F_{jt}$. Any discovered topic of interest was set as a custom topic for monitoring instances of it in the following time periods.

Figure 3 illustrates the quarterly trend of custom topics. The custom topics, such as Air Bags and Engine represent broad topics which at minimum indicate what part is malfunctioning. Sometimes it is difficult to build custom topics with terms indicative of symptoms, because those symptoms might be new, or organizations might not have sufficient domain expertise to identify all possible symptoms. In such case a close examination discovered topics can help identify symptoms. For example, in this case study, when analyzing the 36 months of historic data, the text topic node identified a topic with the key terms *power*, *steer*, *indicator*, and *light*. These key terms were associated with symptoms such as *lock* and *illuminate*. A close examination of complaints revealed that these symptoms were associated with steering locking issues where malfunction indicator light would turn on. This topic was called "power steering locking." In subsequent quarterly analyses, "power steering locking" was used as a custom topic.

*Figure 3 – Topic Trend*

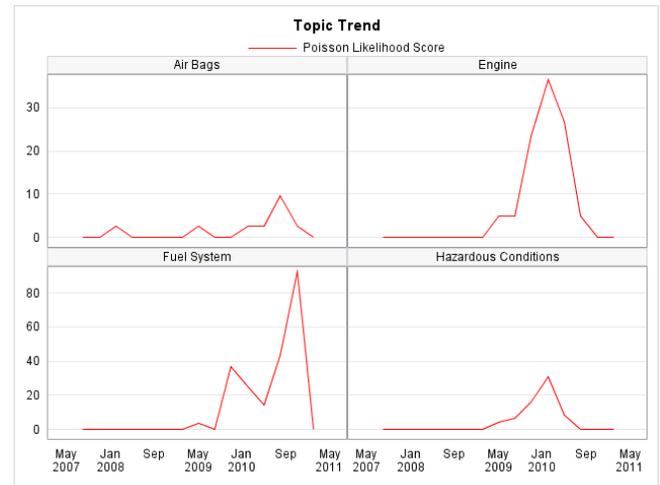

The quarterly trend of "Power Steering Locking" is shown in Figure 4. The figure indicates a sudden upward trend in the Poisson likelihood ratio starting in May 2008. This trend is alarming and indicates a problem that affects a significant number of products in the field.

*Figure 4 – Power Steering Locking Trend*

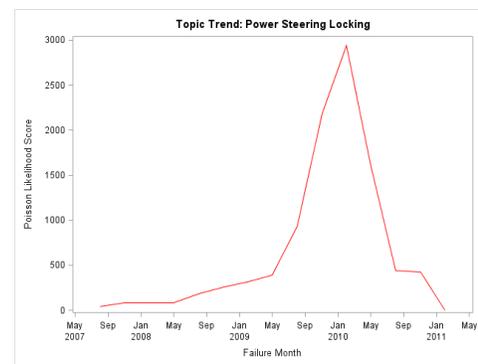

*CONCLUSIONS*

This paper outlines a simple method to detect emerging reliability issues using customer complaints. It uses text mining technology to identify topics and statistical methods to detect emerging issues. The topics, which at minimum consist of product part or sub-system with one or many symptoms, represent very useful information. A manufacturer or a service provider can get insights about symptoms and their prevalence. The method described in this paper cannot be completely automated. As outlined in earlier sections, human input is needed in defining stop lists, synonyms, and custom topics and in interpreting the discovered topics. The human efforts are primarily needed at the project start, when baselines for the topics are established. The log likelihood statistic described in the paper along with some symptom severity statistic, such as average cost of repairs, can help prioritize the issues for further root cause analysis.

## BIOGRAPHIES

Deovrat Kakde
SAS Institute Inc.
World Headquarters
SAS Campus Drive
Cary, NC 27513 USA

e-mail: Dev.Kakde@sas.com

Deovrat Kakde works as a Senior Research Statistician Developer at SAS Institute. He has over fifteen years of experience in various industries such as Manufacturing, Transportation and Software. He holds Master's degree in Quality, Reliability and Operations Research from the Indian Statistical Institute, Kolkata and Bachelor's degree in Production Engineering from Bombay University, India. He is an ASQ Certified Reliability Engineer.

Arin Chaudhuri
SAS Institute Inc.
World Headquarters
SAS Campus Drive
Cary, NC 27513 USA

e-mail: Arin.Chaudhuri@sas.com

Arin Chaudhuri has been working as a Principal Research Statistician Developer at SAS Institute for the past 10 years. He writes code to implement various statistical and numerical algorithms. He holds a Master's and a Bachelor's degree in Statistics from the Indian Statistical Institute, Kolkata.